%% file: main.tex
\documentclass{article}

\usepackage[preprint]{neurips_2026}

\usepackage[utf8]{inputenc} 
\usepackage[T1]{fontenc}    
\usepackage{hyperref}       
\usepackage{url}            
\usepackage{booktabs}       
\usepackage{amsfonts}       
\usepackage{nicefrac}       
\usepackage{microtype}      
\usepackage{xcolor}         
\usepackage{amsmath, amssymb}
\usepackage{enumitem}       
\usepackage{tcolorbox}
\usepackage{longtable}
\usepackage{graphicx}
\usepackage{adjustbox}

\usepackage[capitalize]{cleveref}
\crefname{section}{Sec.}{Secs.}
\Crefname{section}{Section}{Sections}
\Crefname{table}{Table}{Tables}
\crefname{table}{Tab.}{Tabs.}

\title{EditSleuth: A Dataset of Grounded Reasoning Chains for Image-Edit Forensics}

\author{%
  Van-Loc Nguyen\textsuperscript{\rm 1}\thanks{Equal Contribution}, AprilPyone MaungMaung\textsuperscript{\rm 2*}, Minh-Triet Tran\textsuperscript{\rm 1}, Isao Echizen\textsuperscript{\rm 2},\\
  \and
  \textsuperscript{\rm 1}University of Science, Vietnam National University Ho Chi Minh City \\
  \texttt{\{nvloc@selab.,tmtriet@\}hcmus.edu.vn}
  \and
  \textsuperscript{\rm 2}National Institute of Informatics \\
  \texttt{\{pyone, iechizen\}@nii.ac.jp}
}

\begin{document}

\maketitle

\begin{abstract}
Forensic analysis of AI-edited images requires more than binary real-versus-fake prediction: a useful system should localize the edit, identify its semantic type, and ground its decisions in visual evidence. Existing image-forensics datasets typically emphasize detection or localization, while reasoning-supervised vision-language datasets rarely target image manipulation and often rely on LLM-generated rationales whose faithfulness is difficult to verify. We introduce \textbf{EditSleuth}, a dataset of $257{,}725$ image-edit triplets constructed from existing image-editing corpora for grounded image-edit forensic reasoning. Each example includes an edited image, its source image, a binary edit mask, a 12-class edit taxonomy label, a difficulty score, and a six-step reasoning chain. EditSleuth chains are generated deterministically from triplet-grounded upstream artifacts, with each statement tied to a specific computable source of evidence. Our analysis reveals that a naive four-component difficulty formulation suffers from a rank-2 correlation collapse among magnitude features; a simplified three-component formulation substantially increases score dispersion on both Pico-Banana and MagicBrush. Difficulty also varies meaningfully within most edit categories, indicating that the score is not a proxy for edit type. As an initial learning study, we fine-tune Qwen2-VL-2B with LoRA and find that chain-as-target supervision matches a label-only baseline on classification accuracy among parseable answers, while additionally yielding grounded explanatory prose that label-only supervision cannot produce. We release the dataset, the deterministic construction pipeline, and pilot training scripts.
\end{abstract}

\vspace{-.2cm}
\section{Introduction}
\label{sec:intro}
\vspace{-.2cm}


AI-driven image editing has shifted forensic detection from pixel-level splice and copy-move artifacts to coherent, instruction-conditioned edits that may be subtle, semantic, or globally diffuse. Detecting such edits requires more than binary real-versus-fake classification: a useful forensic system should localize the edit, identify its type, and explain the supporting visual evidence. This gap between modern forensic needs and existing dataset supervision motivates our work.

Two adjacent trends provide useful raw material.
First, image-editing datasets such as MagicBrush \citep{zhang2023magicbrush}, InstructPix2Pix \citep{brooks2023instructpix2pix}, UltraEdit \citep{zhao2024ultraedit}, EmuEdit \citep{sheynin2024emu}, and Pico-Banana \citep{qian2025pico} now offer large-scale $(\text{real}, \text{edited}, \text{instruction})$ triplets for training editor models.
Second, reasoning-supervised VLMs show that chain-of-thought supervision can transfer to multimodal tasks \citep{chen2024m3cot, shao2024visual, zhang2025improve}, but their chains are often generated by frontier LLMs that do not directly observe the visual evidence, creating an auditability gap. Neither thread directly addresses forensic detection: editing datasets are used to train editors rather than detectors, and reasoning datasets rarely target image manipulation.

\textbf{EditSleuth} bridges these two threads by treating an image-editing triplet $(I_{\text{real}}, I_{\text{edited}}, t_{\text{instr}})$ as a forensic training example. From the image pair, we recover a binary edit mask; from the instruction or source metadata, we recover the edit category; and from these artifacts, we compose a structured reasoning chain whose grounded claims are computed rather than LLM-synthesized. We implement this idea as a five-stage deterministic pipeline that ingests editing triplets, computes edit masks, scores difficulty, assigns one of 12 canonical categories, and emits a six-step reasoning chain in which every triplet-grounded statement traces to a specific upstream artifact. Applied to Pico-Banana single-turn, the pipeline yields $257{,}725$ fully annotated triplets; we also process MagicBrush dev ($528$ triplets) for held-out cross-instruction-style evaluation.

We characterize EditSleuth and validate its construction through dataset analyses and pilot training. The V2 difficulty formula increases score standard deviation by $+55\%$ on Pico-Banana and $+94\%$ on MagicBrush by correcting a rank-2 collapse among structural, perceptual, and locality-based magnitude terms. Threshold calibration across LAB+SSIM, LAB+LPIPS, and LAB+LPIPS+SSIM shows that threshold shifts track changes in mean combined diff signal, indicating compensation for max-pool inflation rather than signal-specific artifacts. The category-by-difficulty cross-tab further shows that difficulty varies within most categories, with \texttt{geometric} skewing hard ($62\%$) and \texttt{text\_edit} skewing easy ($50\%$). Finally, a matched-budget pilot fine-tuning study finds that chain-target supervision matches label-only classification quality conditional on parseable outputs, while additionally producing grounded prose. However, chain-target extraction recall remains lower than label-only recall and does not close after one additional epoch.


\textbf{Contributions.} (i) A construction pipeline that re-purposes existing image-editing triplets as forensic-detection training data, with grounded reasoning chains composed deterministically from upstream artifacts. (ii) A $12$-category taxonomy of image edits adapted from the Pico-Banana grouping, plus a dataset of $n=257{,}725$ triplets populated through the full pipeline, with parallel MagicBrush dev coverage for held-out evaluation. (iii) Empirical validation of the pipeline's design choices: the rank-2 collapse motivating V2 difficulty, the threshold-calibration observation about max-pool inflation, and the within-category discrimination of the difficulty score. (iv) A pilot fine-tuning study that validates chain-as-target supervision at the classification level and surfaces format conformance as a remaining bottleneck for follow-up work.
\vspace{-.2cm}

\vspace{-.2cm}
\section{Related work}
\label{sec:related-work}
\vspace{-.3cm}
 
EditSleuth lies at the intersection of image-forensics datasets, image-editing datasets, and reasoning-supervised VLM training. We review these threads and use their synthesis to position our contribution.

\textbf{Image forensics datasets.} Image-forensics datasets have developed in two main generations. Early benchmarks for classical manipulations such as splicing and copy-move established the pairing of manipulated images with binary masks localizing edited regions \citep{dong2013casia, hsu2006detecting, wen2016coverage}. These datasets remain important but predate AI-driven editing and provide neither instruction context nor explanations beyond masks. More recent datasets address generative synthesis and editing, benchmarking detectors on AI-generated or AI-edited images \citep{wang2023dire, chang2025antifakeprompt, kang2025legion, jifakexplain, ye2025loki, huang2025sida, xu2025fakeshield, wen2026spot, han2026vrag}. However, most retain a single-label framing, such as real/fake or edit class, without modeling edit localization or the detector's evidential reasoning. EditSleuth fills this gap by combining masks, category labels, and structured reasoning chains that expose the forensic signals behind each decision.

\textbf{Image editing datasets.} Large-scale image-editing datasets pair a source image, an edit instruction, and the edited result. InstructPix2Pix \citep{brooks2023instructpix2pix} introduced synthesized edit pairs using language-model-expanded instructions and text-to-image generation. MagicBrush \citep{zhang2023magicbrush} added human-annotated edits with natural conversational instructions, while later datasets scaled this paradigm with broader generation pipelines and quality controls \citep{zhao2024ultraedit, sheynin2024emu, hui2024hq}. Pico-Banana \citep{qian2025pico} further provides a curated 257K single-turn corpus organized by an edit-type taxonomy. These datasets train editors to map $(I_{\text{real}}, t_{\text{instr}})$ to $I_{\text{edited}}$. EditSleuth re-purposes the same triplets for the reverse task: given $(I_{\text{real}}, I_{\text{edited}}, t_{\text{instr}})$, produce a forensic explanation of the edit.

\textbf{Reasoning-supervised VLM training.} A third line of work studies how VLMs acquire explicit reasoning. LLaVA-style instruction tuning showed that synthetic instruction-response pairs can produce capable multimodal dialogue \citep{liu2023visual}, with later work scaling the approach through larger and more diverse corpora \citep{liu2024llavanext, dai2023instructblip}. Recent efforts further introduce explicit reasoning formats, including multimodal chain-of-thought, faithful multimodal reasoning, and visual chain-of-thought supervision \citep{chen2024m3cot, zhang2023multimodal, li2026faithfulfirstreasoningplanningacting, shao2024visual}. Related work in synthetic-image detection also uses VLMs to generate explanations for AI-generated image forensics \citep{jiang2026fake, guo2025rethinking, huang2025sida, wen2026spot}.

However, two limitations persist. Reasoning chains are often generated by frontier LLMs that do not directly observe the evidence they describe, making fluency easier to guarantee than faithfulness. Their factual claims are also hard to verify at scale without dense annotations. EditSleuth addresses both by composing chains from image-pair-derived artifacts: each triplet-grounded claim traces to a computed upstream value and is auditable end-to-end. Category-level priors are kept separate and marked as category-conditional knowledge.

\textbf{Positioning.} EditSleuth fills the gap between three lines of work.
Forensics datasets provide masks and labels but lack instruction context and reasoning chains.
Editing datasets provide instruction-conditioned triplets but target image editing rather than forensic explanation.
Reasoning-supervised VLM datasets provide chains but often rely on LLM-generated rationales whose faithfulness is difficult to audit.
EditSleuth repurposes image-editing corpora for forensic supervision and adds deterministic, artifact-grounded reasoning chains. To our knowledge, it is the first dataset to combine edited-image triplets with auditable reasoning-chain supervision for forensic VLM training.

\vspace{-.3cm}
\section{The EditSleuth pipeline}
\label{sec:pipeline}
\vspace{-.3cm}

\begin{figure}[t]
  \centering
  \includegraphics[width=\linewidth]{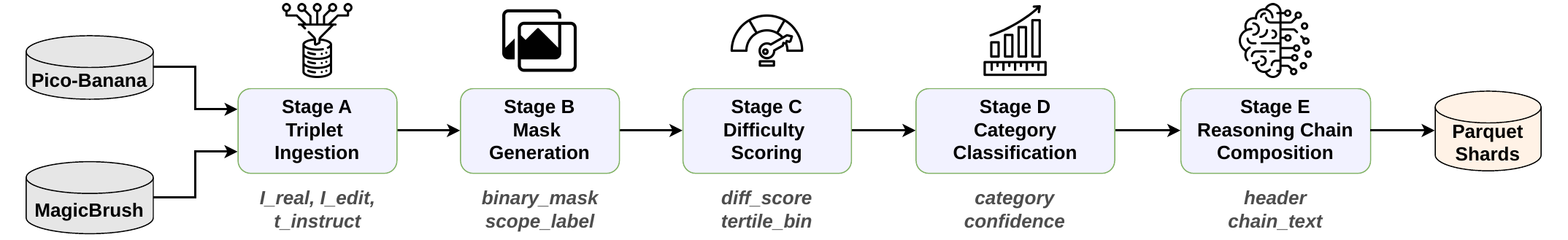}
  \caption{The EditSleuth pipeline. Stage A ingests triplets from source datasets (Pico-Banana~\mbox{\citep{qian2025pico}}, MagicBrush~\citep{zhang2023magicbrush}) into a unified schema. Stages B--E each consume the previous stage's output and produce a per-triplet artifact: a binary edit mask (Stage~B), a difficulty score (Stage~C), a canonical category label (Stage~D), and a six-step reasoning chain (Stage~E).}
  \label{fig:pipeline-overview}
\end{figure}

EditSleuth converts existing image-editing datasets into supervised data for image-edit forensics. Given a triplet $(I_{\text{real}}, I_{\text{edited}}, t_{\text{instr}})$ consisting of a source image, its edited counterpart, and the edit instruction, our five-stage pipeline produces a schema-typed record with a binary edit mask, difficulty score, canonical edit category, and structured reasoning chain for downstream VLM supervision (Figure~\ref{fig:pipeline-overview}).
Each stage is deterministic and emits Parquet records joined by a triplet identifier. The Stage~E reasoning chains are grounded by construction: triplet-specific claims are computed from upstream artifacts, while category-level forensic priors are explicitly marked. This design makes the chains faithful supervision rather than free-form rationales (\S\ref{sec:stage-reasoning}). Stage~A handles dataset ingestion through adapters described in Appendix~\ref{app:adapters}.

\subsection{Stage B: mask generation}
\label{sec:stage-mask}
\vspace{-.2cm}

Stage~B derives a binary edit mask and scope label from each $(I_{\text{real}}, I_{\text{edited}})$ pair, where scope $\in \{\texttt{local}, \texttt{global}, \texttt{ambiguous}, \texttt{alignment\_failed}\}$. Local masks identify spatially bounded edits, while global masks denote image-wide changes. Qualitative mask examples are in Appendix~\ref{app:image-mask-examples}.

\vspace{-.2cm}
\paragraph{Multi-signal differencing.}
We compute three complementary pixel-level difference signals: $L^*a^*b^*$ color distance, structural dissimilarity ($1{-}\mathrm{SSIM}$), and LPIPS perceptual distance \citep{zhang2018unreasonable}. Each signal is normalized to $[0,1]$ by its per-pair $99^{\text{th}}$ percentile, then combined by elementwise maximum. This stack captures chromatic, structural, and perceptual edits, improving MagicBrush dev IoU by 1.3 points over the best two-signal subset (paired bootstrap $p < 10^{-4}$, 95\% CI $[0.0085, 0.0164]$).

\vspace{-.2cm}
\paragraph{Two-path scope routing.}
Given the combined diff map $d \in [0,1]^{H \times W}$, we assign edit scope with a two-path rule. We first route to \texttt{global} if $\mathrm{mean}(d) > \tau$, with $\tau$ calibrated per dataset. Otherwise, we binarize $d$ using Otsu's method \citep{otsu1975threshold}, apply a small morphological opening, and assign scope by mask area: \texttt{global} if $>90\%$, \texttt{local} if $0.5$--$90\%$, and \texttt{ambiguous} if $<0.5\%$. Registration failures during $L^*a^*b^*$ computation are logged as \texttt{alignment\_failed} and propagated downstream.
We calibrate $\tau$ to target a ${\sim}30\%$ global routing rate per dataset ($\tau=0.62$, $30.4\%$ on Pico-Banana). The full calibration in \S\ref{sec:val-threshold} shows that $\tau$ offsets across signal-stack variants track offsets in combined diff means, suggesting that calibration mainly compensates for max-pool inflation rather than signal-specific artifacts.

\vspace{-.2cm}
\subsection{Stage C: difficulty scoring}
\label{sec:stage-difficulty}
\vspace{-.2cm}
 
Stage~C assigns each triplet a scalar difficulty score and tertile bin $\in \{\texttt{easy}, \texttt{medium}, \texttt{hard}\}$ for curriculum sampling and per-difficulty evaluation.

\vspace{-.2cm}
\paragraph{Three-component formula.}
The production scorer (V2) defines difficulty as
\[
  D \;=\; w_S \cdot s_{\text{struct}}
        \;+\; w_C \cdot s_{\text{compact}}
        \;+\; w_I \cdot s_{\text{instr}}
\]

with default weights $(w_S,w_C,w_I)=(0.55,0.25,0.20)$. The components measure edit magnitude, spatial dispersion, and instruction complexity:
\[
s_{\text{struct}} = 1-\mathrm{SSIM}(I_{\text{real}}, I_{\text{edited}}),
\]
\[
s_{\text{compact}} = 1-\mathrm{compactness}(M), \qquad
\mathrm{compactness}(M)=
\sqrt{
\tfrac{|M|}{|\mathrm{bbox}(M)|}
\cdot
\tfrac{|M_{\text{largest cc}}|}{|M|}
},
\]
where $M$ is the Stage~B binary mask, and $s_{\text{instr}}$ is a heuristic score based on instruction length, verb count, conjunction count, and spatial-reference count.
 
The choice of three components rather than V1, four (structural, perceptual, locality, instruction) follows from a correlation- structure observation: the magnitude trio collapses to rank-2 on both datasets (pairwise $|r| \in [0.68, 0.91]$), suppressing the variance of the combined score under weighted summation. V2 retains $s_{\text{struct}}$ as the magnitude representative and adds compactness as a genuinely independent spatial signal, widening the score's standard deviation by $+94\%$ on MagicBrush and $+55\%$ on Pico-Banana relative to the four-component baseline (\S\ref{sec:val-difficulty}).

\vspace{-.2cm}
\paragraph{Tertile binning.}
We compute the empirical $33^{\text{rd}}$ and $66^{\text{th}}$ percentiles of $D$ on each dataset and bin triplets accordingly. Binning is dataset-internal so that \texttt{easy/medium/hard} labels remain meaningful relative to corpus statistics rather than absolute thresholds.

\vspace{-.2cm}
\subsection{Stage D: category classification}
\label{sec:stage-category}
\vspace{-.2cm}

Stage~D assigns each triplet a canonical edit-category label drawn from a 12-category taxonomy: \texttt{object\_addition}, \texttt{object\_removal}, \texttt{object\_replacement}, \texttt{attribute\_change}, \texttt{style\_transfer}, \texttt{photometric}, \texttt{scene\_transformation}, \texttt{background\_change}, \texttt{text\_edit}, \texttt{geometric}, \texttt{human\_centric}, and an explicit \texttt{other} bucket.

\vspace{-.2cm}
\paragraph{Taxonomy choice.}
Our taxonomy adapts the eight-category Pico-Banana grouping \citep{qian2025pico} with three changes. First, we separate \texttt{photometric} edits from \texttt{style\_transfer}, since global tone, grain, and color shifts have different forensic signatures from structural texture rewrites. Second, we add \texttt{human\_centric} for 14 person-specific, identity-preserving operations, including pose, expression, clothing, Funko-Pop, and LEGO-style edits. Third, we keep \texttt{background\_change} distinct from \texttt{scene\_transformation} to capture foreground-preserving background swaps present in MagicBrush but absent from Pico-Banana's taxonomy.

\vspace{-.2cm}
\paragraph{Two-path classification.}
Stage~D assigns edit categories through either source-label mapping or instruction-based rules. For datasets with per-row labels, such as Pico-Banana's \texttt{source\_edit\_type}, we use a hand-curated lookup covering all 35 distinct Pico-Banana labels. For instruction-only datasets, such as MagicBrush, we apply an ordered regex classifier over explicit edit verbs, conversational frames, domain keywords, and geometric operations. Each prediction records its \texttt{source} and confidence; unmapped cases route to \texttt{other} with the original string preserved for audit. Pico-Banana classifies entirely through the label path with zero \texttt{other}, while MagicBrush dev reaches 56\% rule coverage, leaving 44\% as \texttt{other} (\S\ref{sec:limitations}).

\vspace{-.2cm}
\subsection{Stage E: reasoning chain composition}
\label{sec:stage-reasoning}
\vspace{-.2cm}
 
Stage~E generates the per-triplet reasoning target for downstream VLM training: a structured one-line header for filtering and a six-step prose chain used as the supervised output.

\vspace{-.2cm}
\paragraph{The faithfulness model.}
EditSleuth separates two statement types by construction. \emph{Triplet-grounded statements} describe the current example and are sourced from upstream artifacts: the instruction from Stage~A, localization and area from Stage~B, structural-change and compactness scores from Stage~C, and category label and confidence from Stage~D. \emph{Category-level priors} describe typical forensic signatures for the predicted edit type, using one of 12 hand-curated templates. Priors are explicitly hedged (e.g., ``Edits of this type \emph{typically} exhibit\dots'') so models distinguish per-triplet evidence from category-conditional knowledge.

A naive LLM-based approach could produce fluent but unsupported rationales. In contrast, template-driven composition only states claims licensed by upstream artifacts, making numerical and categorical claims auditable end-to-end. Faithfulness is thus artifact-relative: upstream biases, such as residual structural--compactness correlation (\S\ref{sec:val-difficulty}), propagate into the chains and are documented.

\vspace{-.3cm}
\paragraph{Chain structure.}
Each chain contains six numbered steps (${\sim}80$--$150$ words): (1) the quoted edit instruction; (2) spatial localization using a coarse descriptor from \{\texttt{whole\_image}, four quadrants, \texttt{centered}, \texttt{scattered}, \texttt{alignment\_failed}\}; (3) magnitude from structural-change and compactness scores; (4) edit category and classification source; (5) category-level forensic prior; and (6) difficulty bin and score. The fixed order provides uniform supervision, while each step is instantiated from triplet-specific artifacts.
Full per-category examples of reasoning chains appear in Appendix~\ref{app:chain-examples}.

The 12 forensic-prior templates are listed in Appendix~\ref{app:templates}. Each specifies category-relevant forensic cues, such as boundary discontinuities, lighting inconsistencies, inpainting artifacts, global histogram shifts, and font-rendering noise.

\vspace{-.3cm}
\section{Dataset characterization}
\label{sec:characterization}
\vspace{-.3cm}

We characterize EditSleuth by scale and category coverage (\S\ref{sec:char-scale}), mask-derived spatial properties (\S\ref{sec:char-spatial}), and difficulty distribution (\S\ref{sec:char-difficulty}). Results come from one pipeline run on the Pico-Banana single-turn corpus \citep{qian2025pico}, our primary training source, with MagicBrush dev \citep{zhang2023magicbrush} as a smaller held-out comparison.

\vspace{-.2cm}
\subsection{Scale and category coverage}
\label{sec:char-scale}
\vspace{-.2cm}
 
The Pico-Banana derivation yields $257{,}725$ triplets, each with a binary edit mask (Stage~B), difficulty score, and tertile bin (Stage~C), canonical edit category (Stage~D), and six-step reasoning chain (Stage~E). One corrupted source pair was dropped during Stage~B, with no additional downstream loss. The parallel MagicBrush dev split contains $528$ triplets processed through the same pipeline and is used for held-out inspection and cross-dataset analysis.
 
\begin{table}[t]
  \centering
  \small
  \setlength{\tabcolsep}{4pt}
  \caption{Per-category triplet counts and within-category difficulty distributions. The left block reports counts for each source dataset: Pico-Banana uses curated dataset labels, while MagicBrush uses rule-based classification. The right block reports Pico-Banana within-category difficulty distributions, with rows summing to $100\%$. \texttt{background\_change} is reserved for foreground-preserving background swaps, which appear in MagicBrush but not Pico-Banana; categories with zero Pico-Banana coverage have undefined difficulty distributions and are marked ``---''. \texttt{geometric} edits skew \texttt{hard}, while localized \texttt{text\_edit} examples skew \texttt{easy}.}
  \label{tab:cat-joint}
  \begin{tabular}{lrrrr@{\hspace{12pt}}rrr}
    \toprule
    & \multicolumn{2}{c}{Pico-Banana} & \multicolumn{2}{c}{MagicBrush dev} & \multicolumn{3}{c}{Difficulty (Pico-Banana, \%)} \\
    \cmidrule(lr){2-3} \cmidrule(lr){4-5} \cmidrule(lr){6-8}
    Category               & count    & \% & count & \% & easy & medium & hard \\
    \midrule
    object\_addition       & 14{,}187  &  5.5 & 151 & 28.6 & 37.4 & 35.8 & 26.8 \\
    object\_removal        & 15{,}109  &  5.9 &  42 &  8.0 & 37.6 & 36.9 & 25.6 \\
    object\_replacement    & 14{,}547  &  5.6 &  26 &  4.9 & 41.2 & 35.5 & 23.3 \\
    attribute\_change      & 24{,}600  &  9.5 &  49 &  9.3 & 38.3 & 32.4 & 29.3 \\
    style\_transfer        & 42{,}871  & 16.6 &   0 &  0.0 & 31.1 & 36.1 & 32.8 \\
    photometric            & 30{,}188  & 11.7 &   0 &  0.0 & 43.8 & 31.4 & 24.8 \\
    scene\_transformation  & 52{,}691  & 20.4 &   7 &  1.3 & 32.3 & 34.8 & 32.9 \\
    background\_change     &      0    &  0.0 &  11 &  2.1 & ---  & ---  & ---  \\
    text\_edit             & 10{,}688  &  4.1 &   5 &  0.9 & 50.4 & 30.6 & 19.0 \\
    geometric              & 32{,}742  & 12.7 &   2 &  0.4 & 12.8 & 25.2 & 62.0 \\
    human\_centric         & 20{,}102  &  7.8 &   0 &  0.0 & 31.6 & 36.4 & 32.0 \\
    \emph{other}           &      0    &  0.0 & 235 & 44.5 & ---  & ---  & ---  \\
    \midrule
    total                  & 257{,}725 & 100  & 528 & 100  &      &      &      \\
    \bottomrule
  \end{tabular}
\end{table}
 
Table~\ref{tab:cat-joint} reports per-category counts. Pico-Banana covers all 11 non-\texttt{other} categories, with no category above 21\% and the smallest non-zero category, \texttt{text\_edit}, at 4.1\%. Its \texttt{other} bucket is empty because the source taxonomy maps fully to our canonical labels. MagicBrush dev covers a different subset, dominated by object additions and attribute changes, with smaller removal, replacement, and \texttt{background\_change} groups. Its 44.5\% \texttt{other} rate reflects the limits of regex classification on free-text conversational instructions (\S\ref{sec:limitations}).
 
The two corpora play complementary roles. Pico-Banana provides a large, taxonomically curated training source, while MagicBrush serves as a smaller held-out set with naturally phrased instructions for testing generalization to a different instruction style. We therefore do not merge them into a single training pool.

\vspace{-.2cm}
\subsection{Spatial properties of the edits}
\label{sec:char-spatial}
\vspace{-.2cm}
 
Stage~B classifies $30.4\%$ of Pico-Banana triplets as \texttt{global} and $69.6\%$ as \texttt{local}. \texttt{photometric} and \texttt{style\_transfer} edits are global by construction, while \texttt{scene\_transformation} is mostly global because weather, lighting, and seasonal changes affect the full scene. Object-level categories and \texttt{human\_centric} edits are predominantly local, with masks concentrated on the modified region.
 
For local triplets, Stage~E assigns a coarse spatial descriptor: one of four quadrants, \texttt{centered}, or \texttt{scattered}. On Pico-Banana, the distribution is highly skewed: \texttt{centered} covers $89\%$ of local edits, quadrants account for only ${\sim}5\%$ in total ($1.0$--$1.8\%$ each), and \texttt{scattered} covers $5.5\%$. This reflects Pico-Banana's curated, subject-centered imagery rather than a descriptor artifact. We retain the descriptor for faithfulness, while preserving mask area fraction as the more informative spatial signal in Step~2 of the reasoning chain.
 
%

\vspace{-.2cm}
\subsection{Difficulty distribution}
\label{sec:char-difficulty}
\vspace{-.2cm}
 
The Stage~C V2 score on Pico-Banana has a mean of $0.450$ and standard deviation of $0.109$, with tertile cutoffs at $0.406$ and $0.504$. By construction, tertile binning produces $85{,}909$ \texttt{easy}, $85{,}908$ \texttt{medium}, and $85{,}908$ \texttt{hard} triplets. Component statistics show that difficulty is driven mainly by structural change and compactness: $s_{\text{struct}}$ has $\mu=0.446$, $\sigma=0.244$; $s_{\text{compact}}$ has $\mu=0.323$, $\sigma=0.258$; and $s_{\text{instr}}$ has $\mu=0.621$, $\sigma=0.108$.
 
 
Table~\ref{tab:cat-joint} reports within-category difficulty bins. \texttt{geometric} edits, including zoom, outpainting, and object relocation, skew strongly \texttt{hard} ($62.0\%$) because canvas-level transformations increase both structural change and mask dispersion. In contrast, \texttt{text\_edit} skews \texttt{easy} ($50.4\%$), reflecting its typically small, localized changes with modest structural impact.

The other nine categories are more evenly distributed, with no bin exceeding ${\sim}40\%$ within a category. This reflects broad within-category variation: an object addition may be a small accessory or a large scene element, while a photometric edit may range from a subtle tone shift to a strong filter. This spread shows that difficulty is not merely a proxy for category; instead, it provides an additional curriculum axis by discriminating edit magnitude within most categories.
 
We qualitatively validated the bins by inspecting a stratified sample of ${\sim}50$ Pico-Banana triplets. Clear, well-localized edits typically fell in \texttt{easy} or \texttt{medium}, while saturated whole-image masks or visually diffuse edits fell in \texttt{hard}. This pattern held across categories, consistent with the cross-tab evidence that most categories span the difficulty range.

The category-by-difficulty distribution supports two downstream uses. First, curriculum sampling can stratify jointly by category and difficulty, exposing models to both local edits and challenging global transformations throughout training. Second, the reported marginals provide a natural evaluation breakdown: strong performance on \texttt{easy} \texttt{text\_edit} but weak performance on \texttt{hard} \texttt{geometric} implies a different failure mode than uniform performance across the joint distribution.


\begin{figure}
  \centering
  \includegraphics[width=0.5\linewidth]{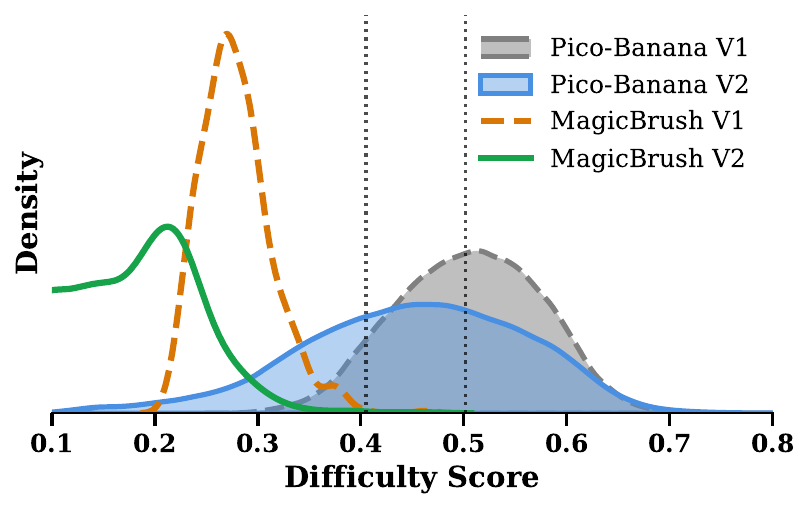}
   \caption{V1 vs V2 difficulty score distributions, side-by-side, on Pico-Banana ($n=257{,}725$) and MagicBrush dev split ($n=528$). V2 widens the standard deviation by $+55\%$ ($\sigma_{V1}=0.070, \sigma_{V2}=0.109$) and produces tertile bins separated by $\sim 1\sigma$ rather than by margins below noise.}
   \label{fig:v1-v2-distribution}
\end{figure}

\vspace{-.3cm}
\section{Experiments}
\label{sec:experiments}
\vspace{-.2cm}

We validate EditSleuth along four axes: difficulty-score ablations against a four-component baseline (\S\ref{sec:val-difficulty}), threshold calibration across signal stacks and datasets (\S\ref{sec:val-threshold}), qualitative inspection of reasoning chains (\S\ref{sec:val-chain-quality}), and a pilot fine-tuning comparison between chain supervision and label-only training (\S\ref{sec:val-pilot}).

\vspace{-.2cm}
\subsection{Difficulty formula ablation}
\label{sec:val-difficulty}
\vspace{-.2cm}

V2 collapses the magnitude trio (structural change, perceptual change, locality) of the four-component baseline (V1) into its single canonical representative ($s_{\text{struct}}$) and adds mask compactness as an independent spatial term (\S\ref{sec:stage-difficulty}).
We score the same Stage~B outputs with both formulas, computing tertile bins separately for each scorer, on Pico-Banana ($n=257{,}725$) and MagicBrush dev ($n=528$).

\vspace{-.2cm}
\paragraph{Variance widening.}
V2 increases score standard deviation by $+94\%$ on MagicBrush ($0.034 \rightarrow 0.066$) and $+55\%$ on Pico-Banana ($0.070 \rightarrow 0.109$). The larger MagicBrush gain reflects stronger V1 rank collapse: magnitude-term correlations are $[0.78,0.91]$ on MagicBrush versus $[0.68,0.85]$ on Pico-Banana.
 

\vspace{-.2cm}
\paragraph{Why the trio collapses.}
The collapse is driven by the strong negative correlation between $s_{\text{loc}} = 1-\mathrm{mask\_area\_frac}$ and the two magnitude terms. Large edits raise $s_{\text{struct}}$ and $s_{\text{perc}}$ but lower $s_{\text{loc}}$ because the mask area grows, causing partial cancellation in the weighted sum. This compresses the combined score to roughly $1/4$--$1/5$ of a single component's variance. V2 avoids this by using one canonical magnitude term and adding compactness, which is more independent of edit magnitude.

One residual issue remains: on Pico-Banana local edits, V2's image-derived components, $s_{\text{struct}}$ and $s_{\text{compact}}$, show a moderate negative correlation ($r=-0.44$). This likely reflects Stage~B mask-quality artifacts, where subtle diffs produce noisier, less compact masks. We discuss this limitation and propose image-pair embedding distance as a future third component in \S\ref{sec:limitations}.

\vspace{-.2cm}
\subsection{Threshold calibration across signal stacks}
\label{sec:val-threshold}
\vspace{-.2cm}
 
Stage~B uses a per-dataset threshold $\tau$ on the mean combined diff signal for Path-1 global routing. We compare three signal stacks---LAB+SSIM, LAB+LPIPS, and LAB+LPIPS+SSIM---and calibrate $\tau$ to produce a ${\sim}30\%$ Path-1 global rate on Pico-Banana. This target matches the share of Pico-Banana labels corresponding to whole-image edits, including style transfer, photometric overlays, scene transformations, and geometric canvas operations.
 
\begin{table}[t]
  \centering
  \small
  \caption{Per-stack calibrated thresholds on Pico-Banana ($n=257{,}725$). Threshold offsets closely track shifts in the mean combined diff signal (cdm.mean), suggesting that $\tau$ primarily compensates for max-pool inflation rather than signal-specific artifacts.}
  \label{tab:threshold-calibration}
  \begin{tabular}{lrrr}
    \toprule
    Signal stack       & cdm.mean & calibrated $\tau$ & Path-1 rate (\%) \\
    \midrule
    LAB + SSIM         & 0.429    & 0.52              & 28.4 \\
    LAB + LPIPS        & 0.465    & 0.56              & 29.1 \\
    LAB + LPIPS + SSIM & 0.530    & 0.62              & 30.4 \\
    \bottomrule
  \end{tabular}
\end{table}

\vspace{-.2cm}
\paragraph{Max-pool inflation.}
Threshold offsets between signal stacks closely track shifts in mean combined diff: $+0.04$ from LAB+SSIM to LAB+LPIPS versus $+0.036$ in cdm.mean, and $+0.06$ from LAB+LPIPS to LAB+LPIPS+SSIM versus $+0.065$. This reflects elementwise max-pooling, which monotonically increases the per-pair diff mean; $\tau$ must rise accordingly to preserve a fixed routing rate. Thus, thresholds do not transfer across stacks directly, but their adjustment is predictable from the cdm.mean shift. Appendix~\ref{app:threshold-tool} provides a retrospective threshold sweep tool.

\vspace{-.2cm}
\subsection{Qualitative chain inspection}
\label{sec:val-chain-quality}
\vspace{-.2cm}
 
We inspect a stratified sample of ${\sim}50$ Pico-Banana triplets across difficulty bins, checking whether triplet-grounded numerical claims match upstream artifacts and whether category-level forensic priors match the visible edit.

We find no faithfulness violations: all numerical claims, including mask area fraction, structural-change score, and difficulty bin, match upstream artifacts by construction. Forensic priors are visually plausible except in two MagicBrush cases where keyword rules misclassified the edit semantics. On Pico-Banana, where categories come from curated labels, priors are always appropriate. Appendix~\ref{app:chain-examples} shows per-category examples.

\vspace{-.2cm}
\subsection{Pilot fine-tuning study}
\label{sec:val-pilot}
\vspace{-.2cm}
 
We run a small pilot to compare chain supervision with label-only training. The study uses one base VLM, one fine-tuning setup, and one held-out evaluation; larger-scale training is left to follow-up work.

\vspace{-.2cm}
\paragraph{Setup.}
We fine-tune Qwen2-VL-2B-Instruct \citep{wang2024qwen2} with LoRA \citep{hu2022lora} ($r=16$, $\alpha=32$; target modules $\{q\_proj, v\_proj\}$ in language layers) on a stratified Pico-Banana subset ($n_{\text{train}}\sim20{,}000$), balanced across 11 non-\texttt{other} categories and three difficulty bins. We compare two variants with identical base model, hyperparameters, and step budget: \emph{chain-target}, which predicts the full Stage~E reasoning chain, and \emph{label-only}, which predicts a JSON $\{\text{category}, \text{scope}, \text{difficulty\_bin}\}$ triple. Evaluation uses MagicBrush dev ($n=528$) for held-out cross-instruction-style generalization; MagicBrush is never seen during training.

\vspace{-.2cm}
\paragraph{Metrics.}
We report category, spatial-descriptor, difficulty-bin, and joint accuracy on the held-out set, counting unextractable predictions as incorrect. We also report field-extraction recall, i.e., the fraction of generations from which each field can be parsed. For the chain-target model, we extract category, spatial descriptor, and difficulty bin from steps 4, 2, and 6, respectively; for the label-only model, we parse all fields directly from JSON.

\vspace{-.2cm}
\paragraph{Results.}
Table~\ref{tab:pilot-results} compares both training configurations on MagicBrush dev across the four evaluation metrics.
 
\begin{table}[t]
  \centering
  \small
  \caption{Pilot fine-tuning results on MagicBrush dev ($n=528$).}
  \label{tab:pilot-results}
  \begin{tabular}{lrrrr}
    \toprule
    & \multicolumn{2}{c}{Chain target} & \multicolumn{2}{c}{Label-only} \\
    \cmidrule(lr){2-3} \cmidrule(lr){4-5}
    Field & accuracy & extracted & accuracy & extracted \\
    \midrule
    category               & 22.5\% & 67.4\% & 37.1\% & 93.6\% \\
    spatial descriptor     & 29.7\% & 79.9\% & 42.2\% & 100\%  \\
    difficulty bin         & 28.0\% & 68.6\% & 33.1\% & 100\%  \\
    \emph{joint (all 3)}   &  1.9\% & ---    &  3.4\% & ---    \\
    \bottomrule
  \end{tabular}
\end{table}
 
Label-only achieves higher raw per-field accuracy, mainly because its extraction recall is much higher (${\sim}94$--$100\%$ vs.\ ${\sim}67$--$80\%$ for chain-target). Conditional on successful extraction, chain-target obtains $33.4\%$/$37.2\%$/$40.8\%$ on category/spatial/bin, compared with label-only's $39.7\%$/$42.2\%$/$33.1\%$. Thus, chain-target matches or exceeds label-only on two of three fields once parseability is controlled for; the raw gap is driven primarily by less frequent parseable outputs rather than weaker classification.
 
Absolute accuracies are limited by noisy MagicBrush dev labels: the set lacks curated category annotations, so ``ground truth'' comes from the same rule classifier with a $44\%$ fallback rate (\S\ref{sec:char-scale}). Manual inspection reveals several questionable label assignments. Thus, the relative model comparison is more reliable than the absolute accuracy values.

\vspace{-.2cm}
\paragraph{Effect of additional training.}
A two-epoch re-run with the same hyperparameters does not close the recall gap. Label-only improves modestly (category $37.1\rightarrow39.6\%$, bin $33.1\rightarrow34.7\%$) with recall near $100\%$, while chain-target regresses: spatial extraction recall drops $79.9\rightarrow67.0\%$, and conditional spatial/bin accuracy falls from $37.2\rightarrow32.5\%$ and $40.8\rightarrow34.4\%$. Six-step chain format conformance is therefore harder to learn than JSON output, and the 2B LoRA model appears to plateau below label-only recall.
 

\vspace{-.2cm}
\paragraph{Interpretation.}
When outputs are parseable, chain-target supervision matches label-only training on category, spatial, and difficulty classification, indicating no loss in underlying prediction quality. Its added benefit is grounded prose: all chain-target generations include evidence such as mask area, structural-change score, and category-level forensic priors, which label-only training cannot produce. The main limitation is format conformance, which does not improve with one additional epoch at this scale and likely requires a larger model, more diverse data, or explicit conformance training.

\vspace{-.3cm}
\section{Limitations and broader impact}
\label{sec:limitations}
\vspace{-.2cm}

\paragraph{Limitations.}
We flag six limitations of the dataset and the pilot.
\begin{itemize}[leftmargin=*,topsep=2pt,itemsep=2pt]
  \item \emph{Residual mask-quality coupling.} On Pico-Banana local edits, V2's image-derived components remain moderately anti-correlated ($r=-0.44$; \S\ref{sec:val-difficulty}) because subtle edits yield noisier, less compact masks. Future work may replace compactness with image-pair embedding distance.
  \item \emph{Rule-coverage gap on MagicBrush.} Stage~D assigns $44\%$ of MagicBrush dev triplets to \texttt{other} (\S\ref{sec:char-scale}), reflecting the limits of regex rules on conversational instructions. We therefore use MagicBrush as a held-out evaluation rather than training data.
  \item \emph{Single-source training corpus.} Pico-Banana is generated by one text-conditioned editor, so Stage~B masks reflect that editor's artifact statistics. Models trained on EditSleuth may transfer imperfectly to edits from other generators.
  \item \emph{Artifact-relative faithfulness.} Chains are auditable against upstream masks, difficulty scores, and category labels, but these artifacts are themselves outputs of computational stages. A small human-annotated audit set is a natural next step.
  \item \emph{Pilot scope.} The fine-tuning pilot (\S\ref{sec:val-pilot}) uses one 2B VLM with LoRA, a ${\sim}20$K stratified subset, and MagicBrush dev evaluation. A two-epoch follow-up does not close the chain-target recall gap, suggesting that better format conformance likely requires a larger model, more diverse data, or explicit conformance training.
  \item \emph{Unverified inference-time chain faithfulness.} The pilot model learns to emit six-step chains, including numerical fields, but does not compute those fields from the underlying artifacts. Thus, generated triplet-grounded statements are model estimates rather than verified quantities; measuring this drift remains open.
\end{itemize}

\vspace{-.2cm}
\paragraph{Broader impact.}
EditSleuth has dual-use implications. Positively, it can support forensic detectors for image authentication in journalism, legal evidence, and content moderation; its grounded reasoning chains improve auditability relative to black-box classifiers. Negatively, such detectors could suppress legitimate creative work or intensify an arms race in which generators learn to avoid the forensic cues encoded in our templates. We therefore release EditSleuth under a research-use license and recommend human-in-the-loop deployment, especially in high-stakes settings.

\vspace{-.3cm}
\section{Conclusion}
\label{sec:conclusion}
\vspace{-.2cm}
 
We introduced EditSleuth, a dataset and pipeline that converts image-editing triplets into forensic supervision with edit masks, difficulty scores, category labels, and deterministic six-step reasoning chains grounded in upstream artifacts.
Three construction findings stand out. First, V2 increases difficulty-score standard deviation by $+55\%$ on Pico-Banana and $+94\%$ on MagicBrush by replacing a rank-collapsed magnitude trio with one canonical magnitude term plus compactness (\S\ref{sec:val-difficulty}). Second, threshold offsets across signal stacks track combined-diff means, showing that calibration mainly corrects max-pool inflation rather than signal-specific quirks (\S\ref{sec:val-threshold}). Third, the category-by-difficulty cross-tab shows that difficulty varies within most categories, with \texttt{geometric} skewing hard ($62\%$) and \texttt{text\_edit} skewing easy ($50\%$) (\S\ref{sec:char-difficulty}).
Two follow-up directions remain. First, the pilot should be scaled beyond one base VLM, one LoRA setup, and one held-out evaluation, with emphasis on closing the chain-target extraction-recall gap (\S\ref{sec:val-pilot}). Second, a small human-annotated audit set is needed to evaluate upstream artifact correctness and move the faithfulness guarantee from artifact-relative to ground-truth-relative. We release the dataset, construction pipeline, and pilot training scripts under a research-use license.

\medskip

{
\small
\bibliographystyle{unsrtnat}
\bibliography{refs}
}


\include{appendix}


\end{document}

%% file: appendix.tex
\appendix

\section{Stage A: source-dataset adapters}
\label{app:adapters}

Stage~A ingests source-dataset records into the canonical \texttt{EditTriplet} schema. Each adapter handles the source dataset's format-specific quirks; the rest of the pipeline operates on a single uniform schema thereafter. We ship two adapters with the initial release.

\paragraph{Pico-Banana adapter.}
Source format is JSONL with one record per edit, each containing
fields for the input image (\texttt{local\_input\_image} or
\texttt{open\_image\_input\_url}), the edited image
(\texttt{output\_image}), the natural-language instruction
(\texttt{text}), and a source-curated edit category
(\texttt{edit\_type}, e.g.\ ``Remove an existing object''). The
adapter:
\begin{itemize}[leftmargin=*,topsep=2pt,itemsep=2pt]
  \item Constructs \texttt{triplet\_id} as
    \texttt{picobanana\_\{stem\}} where \texttt{stem} is the sanitized
    filename stem from \texttt{output\_image}. We preserve the full
    stem rather than parsing a numeric index, because not all
    Pico-Banana filenames are numeric (e.g., \texttt{kewsee\_retry1.png}),
    and the integer-only parse produced ${\sim}22\%$ duplicate IDs on
    the full release.
  \item Resolves \texttt{real\_path} from \texttt{local\_input\_image}
    under \texttt{dataset\_root}; falls back to downloading
    \texttt{open\_image\_input\_url} into a cache directory if the
    local file is missing and \texttt{download\_missing=True}.
  \item Sets \texttt{provided\_mask\_path = None} (Pico-Banana ships
    no ground-truth masks; Stage B synthesizes them).
  \item Preserves the source \texttt{edit\_type} string in metadata.
    Stage~D consumes this for category classification rather than
    using it directly as the canonical label.
\end{itemize}

\paragraph{MagicBrush adapter.}
Source format is the HuggingFace \texttt{osunlp/MagicBrush} dataset,
with one row per editing turn. Each row contains \texttt{source\_img},
\texttt{target\_img}, \texttt{mask\_img} (a ground-truth manipulation
mask), \texttt{instruction}, \texttt{img\_id} (the COCO image id), and
\texttt{turn\_index} (1-based). The adapter:
\begin{itemize}[leftmargin=*,topsep=2pt,itemsep=2pt]
  \item Constructs \texttt{triplet\_id} as
    \texttt{magicbrush\_\{split\}\_\{img\_id\}\_t\{turn:02d\}}. The
    split tag prevents train/dev collisions if ingested parquets are
    later merged.
  \item Materializes \texttt{source\_img}, \texttt{target\_img}, and
    \texttt{mask\_img} as PNGs under the output root and records
    their paths.
  \item Decodes MagicBrush's mask encoding correctly: the
    \texttt{mask\_img} field is an RGB image where the edited region
    is painted pure black ($[0,0,0]$) over the target image's color
    content. The naive ``$\mathrm{convert(L)} > 127$'' threshold
    inverts the mask. The adapter checks for all-or-mostly-black
    pixels pattern and inverts as needed.
  \item Sets \texttt{metadata[``source\_is\_authentic''] = True}
    only for \texttt{turn\_index == 1}; for turns $\geq 2$ the
    ``source'' is itself a previously edited image. Downstream code
    requiring pristine real images filters on this flag, or sets
    \texttt{single\_turn\_only=True} at ingest time.
\end{itemize}

\paragraph{Adapter contract.}
Adding a new source dataset requires implementing a single class
inheriting from \texttt{IngestionAdapter} (\texttt{adapters/base.py})
and registering it in the Hydra config. The base class handles
output-parquet writing, deterministic ID generation, and image
materialization; subclasses implement only the source-specific
record-shape parsing.

\section{Forensic-prior templates}
\label{app:templates}

The Stage~E reasoning chain includes one category-level forensic
prior per triplet (Step~5), drawn from the table below by the
classified edit category. Templates are versioned
(\texttt{TEMPLATE\_VERSION = "v1.0"}); the version is recorded in
each chain's structured header for downstream auditability. All
templates are phrased as category-level facts (``edits of this type
typically exhibit\dots'') rather than per-triplet observations, so
the model learns to distinguish category-conditional knowledge from
per-triplet evidence.

\begin{table}[t]
  \centering
  \footnotesize
  \caption{Per-category forensic-signature priors. Each template
    is one or two sentences naming concrete forensic features that
    a detector would plausibly attend to that category.}
  \label{tab:forensic-priors}
  \begin{tabular}{p{3.0cm} p{10cm}}
    \toprule
    Category & Forensic-signature template \\
    \midrule
    \texttt{object\_addition} &
      Boundary discontinuities at the edge of the inserted region,
      and lighting or shadow inconsistencies between the new object
      and its surrounding scene. \\
    \addlinespace[2pt]
    \texttt{object\_removal} &
      Inpainting artifacts where the removed object used to be, such
      as blurred or repeated texture patches that disagree with the
      surrounding context. \\
    \addlinespace[2pt]
    \texttt{object\_replacement} &
      Boundary mismatches at the silhouette of the new object, plus
      scale or perspective inconsistencies if the replacement does
      not match the original object's geometry. \\
    \addlinespace[2pt]
    \texttt{attribute\_change} &
      Color or texture discontinuities along the object's boundary
      where the edited region meets its preserved surroundings, often
      without changes elsewhere in the image. \\
    \addlinespace[2pt]
    \texttt{style\_transfer} &
      Global texture and brushstroke patterns are inconsistent with
      natural photography, applied uniformly across the image
      regardless of original content. \\
    \addlinespace[2pt]
    \texttt{photometric} &
      A global histogram shift or noise overlay applied uniformly to
      all pixels, with the underlying image content semantically
      unchanged from the original. \\
    \addlinespace[2pt]
    \texttt{scene\_transformation} &
      Globally consistent changes in lighting, color temperature, or
      weather effects that affect the whole scene coherently rather
      than any single object. \\
    \addlinespace[2pt]
    \texttt{background\_change} &
      A sharp transition between a preserved foreground subject and
      a newly-introduced background, sometimes with mismatched
      lighting or perspective at the boundary. \\
    \addlinespace[2pt]
    \texttt{text\_edit} &
      Font or rendering artifacts in the modified text region ---
      inconsistent letter spacing, mismatched typefaces, or rendering
      noise distinct from the original photographic text. \\
    \addlinespace[2pt]
    \texttt{geometric} &
      Canvas-level transformations such as cropped boundaries,
      scaled content, or extrapolated regions outside the original
      frame, rather than localized object edits. \\
    \addlinespace[2pt]
    \texttt{human\_centric} &
      Subject-localized stylization or attribute changes confined
      to a person, with the surrounding scene preserved;
      identity-preserving transformations often introduce
      distinctive rendering artifacts around the face and hair. \\
    \addlinespace[2pt]
    \emph{other} &
      Edit characteristics depend on the specific operation; without
      a confirmed category, look broadly for any local boundary
      discontinuities or global statistical shifts. \\
    \bottomrule
  \end{tabular}
\end{table}

\section{Retrospective threshold-sweep tool}
\label{app:threshold-tool}

Stage~B's two-path scope routing depends on a threshold $\tau$ on
the combined diff signal's mean, calibrated per dataset to target
a $\sim 30\%$ Path-1 global routing rate (\S\ref{sec:val-threshold}).
We provide \texttt{scripts/sweep\_global\_threshold.py} to surface
this calibration without re-running Stage~B from scratch.

\paragraph{What it computes.}
For each candidate threshold $\tau' \in T_{\text{candidates}}$, the
tool computes the Path-1 global rate that would result from setting
\texttt{global\_mean\_threshold = $\tau'$}. The computation is exact
because Stage~B records each triplet's
\texttt{combined\_diff\_mean} alongside its scope label; recovering
the Path-1 rate at any other threshold is a simple comparison.

What the tool does \emph{not} simulate is Path~2 (the post-Otsu
area-based promotion to \texttt{global}). Path~2 requires the
binarized mask, which is not retained in Stage~B's per-triplet
record. The reported Path-1 rate is therefore a lower bound on the
resulting total global rate; for typical thresholds in our datasets
the post-Otsu contribution is small (${<}1$ percentage point).

\paragraph{Usage.}
\begin{verbatim}
uv run python scripts/sweep_global_threshold.py \
    paths.masks_parquet=${output_root}/masks/pico_banana.parquet \
    candidate_thresholds=[0.50,0.52,0.54,0.56,0.58,0.60,0.62,0.64]
\end{verbatim}

The output is a CSV report with columns
\texttt{(threshold, path1\_global\_rate, n\_total)}. The user picks
the $\tau$ that lands closest to the desired routing rate. On
Pico-Banana with the LAB+LPIPS+SSIM stack, $\tau = 0.62$ produces a
30.4\% Path-1 rate; this is the calibrated threshold reported in
Table~\ref{tab:threshold-calibration}.

\section{Per-category chain examples}
\label{app:chain-examples}

We show one example reasoning chain per canonical category. Each
example is a Stage~E output verbatim, with the triplet's instruction
quoted in Step~1. The structured header on the first line records
category, scope, difficulty bin, and the source of the category
classification. Steps~1--4 and Step~6 are triplet-grounded; Step~5
is the category-level forensic prior from
Appendix~\ref{app:templates}.

The chains are short ($\sim$80--150 words each); we present them
without surrounding prose, so the format and the within-step
variation across categories are directly comparable.

\newcommand{\chainexample}[2]{%
  \begin{tcolorbox}[colback=gray!5,colframe=black!30,boxsep=2pt,
                    title=#1,fonttitle=\small\ttfamily]
    \footnotesize
    #2
  \end{tcolorbox}%
  \vspace{4pt}%
}

\chainexample{object\_addition}{%
\texttt{[category=object\_addition, scope=local, difficulty=easy, source=rule\_based]}\\[3pt]
1. The edit instruction states: ``add a polar bear''.\\
2. The mask of changed pixels covers roughly 12\% of the image and is concentrated in the lower-left region.\\
3. Structural change relative to the original is minor (SSIM-based score = 0.09), and the edit region is well-concentrated in a single coherent region.\\
4. The edit is classified as object\_addition, inferred from the instruction text via a rule-based keyword match (confidence 0.80).\\
5. Edits of this type typically exhibit boundary discontinuities at the edge of the inserted region, and lighting or shadow inconsistencies between the new object and its surrounding scene.\\
6. Overall, this triplet is easier than average to detect, given clear local geometry and a low-complexity instruction (difficulty score = 0.11, instruction complexity = 0.12).%
}

\chainexample{object\_removal}{%
\texttt{[category=object\_removal, scope=local, difficulty=medium, source=rule\_based]}\\[3pt]
1. The edit instruction states: ``get rid of the framed pictures''.\\
2. The mask of changed pixels covers roughly 17\% of the image and is centered in the image.\\
3. Structural change relative to the original is minor (SSIM-based score = 0.15), and the edit region is moderately concentrated.\\
4. The edit is classified as object\_removal, inferred from the instruction text via a rule-based keyword match (confidence 0.85).\\
5. Edits of this type typically exhibit inpainting artifacts where the removed object used to be, such as blurred or repeated texture patches that disagree with the surrounding context.\\
6. Overall, this triplet is of moderate detection difficulty (difficulty score = 0.21, instruction complexity = 0.06).%
}

\chainexample{object\_replacement}{%
\texttt{[category=object\_replacement, scope=local, difficulty=hard, source=rule\_based]}\\[3pt]
1. The edit instruction states: ``replace the stuffed animals with a pillow.''.\\
2. The mask of changed pixels covers roughly 40\% of the image and is centered in the image.\\
3. Structural change relative to the original is moderate (SSIM-based score = 0.32), and the edit region is well-concentrated in a single coherent region.\\
4. The edit is classified as object\_replacement, inferred from the instruction text via a rule-based keyword match (confidence 0.85).\\
5. Edits of this type typically exhibit boundary mismatches at the silhouette of the new object, plus scale or perspective inconsistencies if the replacement does not match the original object's geometry.\\
6. Overall, this triplet is harder than average to detect, given diffuse geometry or a high-complexity instruction (difficulty score = 0.28, instruction complexity = 0.15).%
}

\chainexample{attribute\_change}{%
\texttt{[category=attribute\_change, scope=local, difficulty=easy, source=rule\_based]}\\[3pt]
1. The edit instruction states: ``let the apples be changed to orange slices''.\\
2. The mask of changed pixels covers roughly 13\% of the image and is concentrated in the lower-right region.\\
3. Structural change relative to the original is minor (SSIM-based score = 0.12), and the edit region is well-concentrated in a single coherent region.\\x
4. The edit is classified as attribute\_change, inferred from the instruction text via a rule-based keyword match (confidence 0.80).\\
5. Edits of this type typically exhibit color or texture discontinuities along the object's boundary where the edited region meets its preserved surroundings, often without changes elsewhere in the image.\\
6. Overall, this triplet is easier than average to detect, given clear local geometry and a low-complexity instruction (difficulty score = 0.10, instruction complexity = 0.08).%
}

\chainexample{style\_transfer}{%
\texttt{[category=style\_transfer, scope=global, difficulty=hard, source=dataset\_label]}\\[3pt]
1. The edit instruction states: ``enhance the image to a modern aesthetic by applying a vibrant, high-contrast color grade with crisp details, brightening the overall scene, and subtly smoothing any visible wear or rust on the bridge.''.\\
2. The mask of changed pixels covers roughly 100\% of the image and spans the entire image.\\
3. Structural change relative to the original is substantial (SSIM-based score = 0.80), and the edit region is well-concentrated in a single coherent region.\\
4. The edit is classified as style\_transfer, based on the dataset's curated edit-type label.\\
5. Edits of this type typically exhibit global texture and brushstroke patterns inconsistent with natural photography, applied uniformly across the image regardless of original content.\\
6. Overall, this triplet is harder than average to detect, given diffuse geometry or a high-complexity instruction (difficulty score = 0.56, instruction complexity = 0.58).%
}

\chainexample{photometric}{%
\texttt{[category=photometric, scope=global, difficulty=medium, source=dataset\_label]}\\[3pt]
1. The edit instruction states: ``colorize the black and white image realistically, depicting natural skin tones, jungle foliage, and gear colors, then subtly shift the overall color temperature towards a cooler tone.''.\\
2. The mask of changed pixels covers roughly 100\% of the image and spans the entire image.\\
3. Structural change relative to the original is substantial (SSIM-based score = 0.64), and the edit region is well-concentrated in a single coherent region.\\
4. The edit is classified as photometric, based on the dataset's curated edit-type label.\\
5. Edits of this type typically exhibit a global histogram shift or noise overlay applied uniformly to all pixels, with the underlying image content semantically unchanged from the original.\\
6. Overall, this triplet is of moderate detection difficulty (difficulty score = 0.48, instruction complexity = 0.67).%
}

\chainexample{scene\_transformation}{%
\texttt{[category=scene\_transformation, scope=local, difficulty=medium, source=rule\_based]}\\[3pt]
1. The edit instruction states: ``let the cabinets be made of dark wood''.\\
2. The mask of changed pixels covers roughly 13\% of the image and is centered in the image.\\
3. Structural change relative to the original is minor (SSIM-based score = 0.10), and the edit region is moderately concentrated.\\
4. The edit is classified as scene\_transformation, inferred from the instruction text via a rule-based keyword match (confidence 0.65).\\
5. Edits of this type typically exhibit globally consistent changes in lighting, color temperature, or weather effects that affect the whole scene coherently rather than any single object.\\
6. Overall, this triplet is of moderate detection difficulty (difficulty score = 0.19, instruction complexity = 0.08).%
}

\chainexample{background\_change}{%
\texttt{[category=background\_change, scope=local, difficulty=easy, source=rule\_based]}\\[3pt]
 1. The edit instruction states: ``it should be a mountain in the background.''.\\
2. The mask of changed pixels covers roughly 10\% of the image and is concentrated in the lower-right region.\\
3. Structural change relative to the original is minor (SSIM-based score = 0.07), and the edit region is moderately concentrated.\\
4. The edit is classified as background\_change, inferred from the instruction text via a rule-based keyword match (confidence 0.75).\\
5. Edits of this type typically exhibit a sharp transition between a preserved foreground subject and a newly-introduced background, sometimes with mismatched lighting or perspective at the boundary.\\
6. Overall, this triplet is easier than average to detect, given clear local geometry and a low-complexity instruction (difficulty score = 0.13, instruction complexity = 0.08).%
}

\chainexample{text\_edit}{%
\texttt{[category=text\_edit, scope=local, difficulty=hard, source=rule\_based]}\\[3pt]
 1. The edit instruction states: ``change the text on the parking meter to say ``NO''.''.\\
2. The mask of changed pixels covers roughly 8\% of the image and is centered in the image.\\
3. Structural change relative to the original is minor (SSIM-based score = 0.02), and the edit region is diffuse or split across multiple sub-regions.\\
4. The edit is classified as text\_edit, inferred from the instruction text via a rule-based keyword match (confidence 0.70).\\
5. Edits of this type typically exhibit font or rendering artifacts in the modified text region — inconsistent letter spacing, mismatched typefaces, or rendering noise distinct from the original photographic text.\\
6. Overall, this triplet is harder than average to detect, given diffuse geometry or a high-complexity instruction (difficulty score = 0.24, instruction complexity = 0.18).%
}

\chainexample{geometric}{%
\texttt{[category=geometric, scope=local, difficulty=hard, source=rule\_based]}\\[3pt]
1. The edit instruction states: ``make the piece of paper hanging on the wall a mirror''.\\
2. The mask of changed pixels covers roughly 14\% of the image and is concentrated in the upper-left region.\\
3. Structural change relative to the original is minor (SSIM-based score = 0.08), and the edit region is diffuse or split across multiple sub-regions.\\
4. The edit is classified as geometric, inferred from the instruction text via a rule-based keyword match (confidence 0.85).\\
5. Edits of this type typically exhibit canvas-level transformations such as cropped boundaries, scaled content, or extrapolated regions outside the original frame, rather than localized object edits.\\
6. Overall, this triplet is harder than average to detect, given diffuse geometry or a high-complexity instruction (difficulty score = 0.23, instruction complexity = 0.19).%
}

\chainexample{human\_centric}{%
\texttt{[category=human\_centric, scope=global, difficulty=hard, source=dataset\_label]}\\[3pt]
1. The edit instruction states: ``transform the main subject (the person playing the flute) into a detailed, expressive black ink line-art sketch, utilizing varied line weights to highlight facial features, the texture of the cap.''.\\
2. The mask of changed pixels covers roughly 100\% of the image and spans the entire image.\\
3. Structural change relative to the original is substantial (SSIM-based score = 0.72), and the edit region is well-concentrated in a single coherent region.\\
4. The edit is classified as human\_centric, based on the dataset's curated edit-type label.\\
5. Edits of this type typically exhibit subject-localized stylization or attribute changes confined to a person, with the surrounding scene preserved; identity-preserving transformations often introduce distinctive rendering artifacts around the face and hair.\\
6. Overall, this triplet is harder than average to detect, given diffuse geometry or a high-complexity instruction (difficulty score = 0.51, instruction complexity = 0.58).%
}

\chainexample{other}{%
\texttt{[category=other, scope=local, difficulty=medium, source=fallback]}\\[3pt]
 1. The edit instruction states: ``have there be a basket of fruit on the counter.''.\\
2. The mask of changed pixels covers roughly 8\% of the image and is centered in the image.\\
3. Structural change relative to the original is minor (SSIM-based score = 0.06), and the edit region is moderately concentrated.\\
4. The edit is classified as other, could not be determined confidently from the available signals; treated as an unspecified edit type.\\
5. Edits of this type typically exhibit edit characteristics that depend on the specific operation; without a confirmed category, look broadly for any local boundary discontinuities or global statistical shifts.\\
6. Overall, this triplet is of moderate detection difficulty (difficulty score = 0.16, instruction complexity = 0.10).%
}




\section{Qualitative mask examples}
\label{app:image-mask-examples}

\begin{figure}
    \centering
    \includegraphics[width=\linewidth]{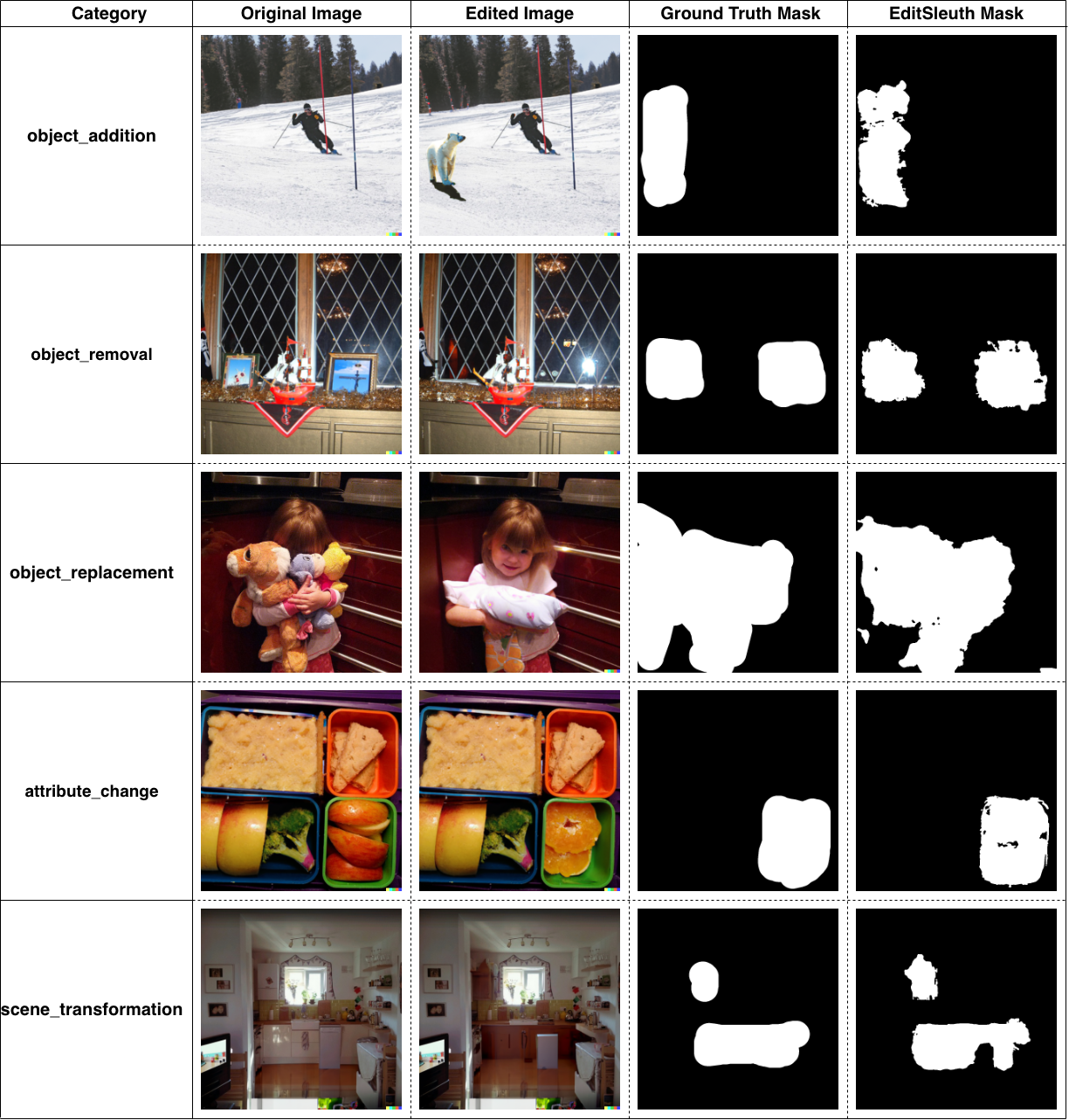}
    \caption{Examples of the original images, edited images, ground-truth masks, and generated masks from EditSleuth for categories \texttt{object\_addition}, \texttt{object\_removal}, \texttt{object\_replacement}, \texttt{attribute\_change}, and \texttt{scene\_transformation}.}
    \label{fig:image-mask-examples-1}
\end{figure}

\begin{figure}
    \centering
    \includegraphics[width=\linewidth]{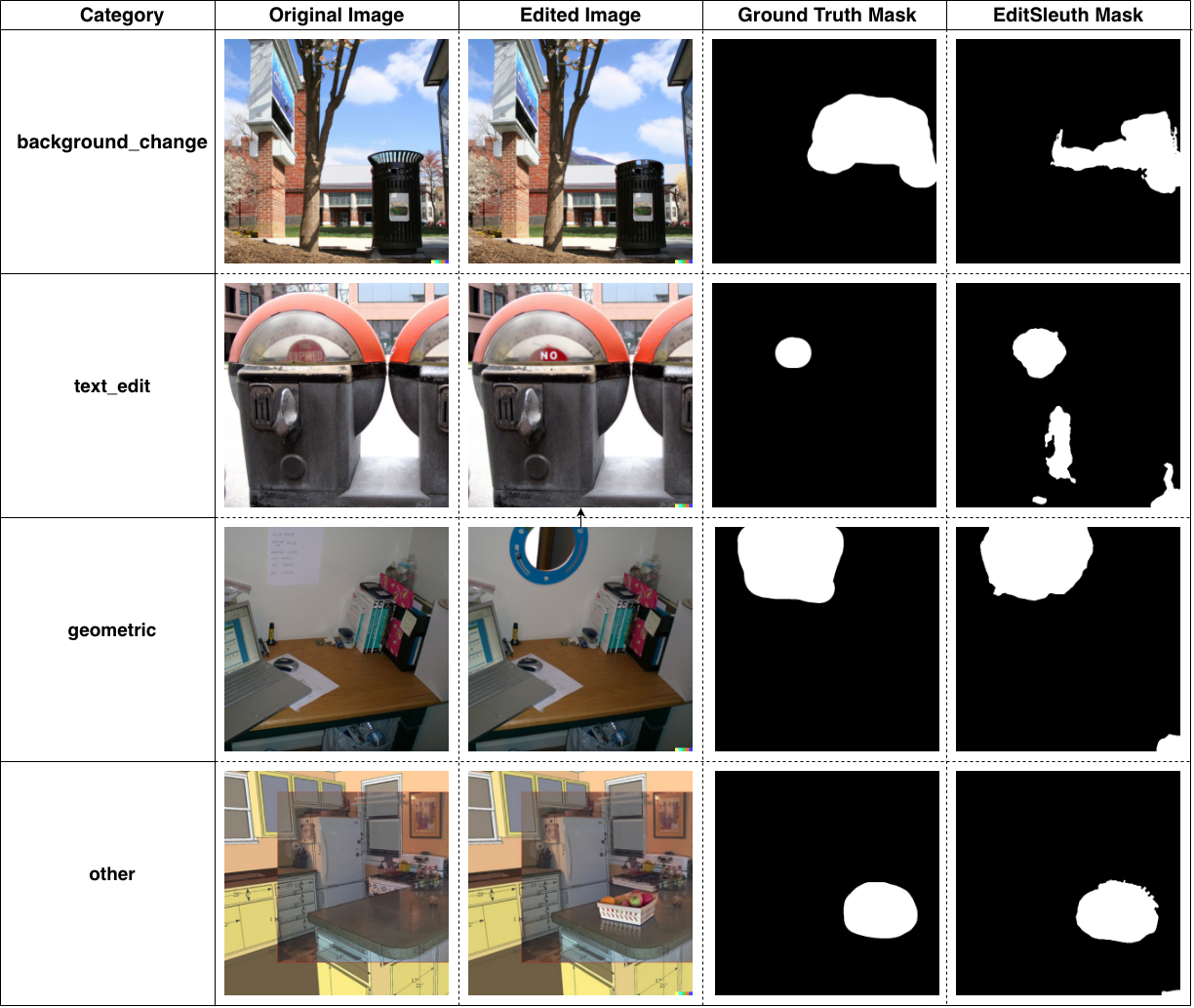}
    \caption{Examples of the original images, edited images, ground-truth masks, and generated masks from EditSleuth for categories \texttt{background\_change}, \texttt{text\_edit}, \texttt{geometric}, and \texttt{other}.}
    \label{fig:image-mask-examples-2}
\end{figure}

Figures~\ref{fig:image-mask-examples-1} and \ref{fig:image-mask-examples-2} compare EditSleuth's Stage~B masks against MagicBrush's
ground-truth manipulation masks across nine representative edit categories. EditSleuth's masks consistently localize the edited region, including multi-component edits (e.g., the two removed picture frames in \texttt{object\_removal} and the two-region \texttt{scene\_transformation}). Their boundaries are morphologically rougher than the ground truth's: derived from
thresholded pixel-difference signals rather than learned segmentation, EditSleuth masks follow the local texture of the
diff signal and exhibit irregular outlines and occasional small false-positives in texture-heavy regions (most visibly in the
\texttt{text\_edit} row).


